\documentclass[conference]{IEEEtran}
\IEEEoverridecommandlockouts
\usepackage{cite}
\usepackage{amsmath,amssymb,amsfonts}
\usepackage[colorlinks,linkcolor=red]{hyperref}
\usepackage{algorithmic}
\usepackage{graphicx}
\usepackage{textcomp}
\usepackage{xcolor}
\usepackage{balance}
\usepackage{geometry}
\usepackage{multicol} 
\usepackage{fancyhdr}

\def\BibTeX{{\rm B\kern-.05em{\sc i\kern-.025em b}\kern-.08em
    T\kern-.1667em\lower.7ex\hbox{E}\kern-.125emX}}
\begin{document}

\title{FewSAR: A Few-shot SAR Image Classification Benchmark\\
}

\makeatletter
\newcommand{\linebreakand}{%
  \end{@IEEEauthorhalign}
  \hfill\mbox{}\par
  \mbox{}\hfill\begin{@IEEEauthorhalign}
}
\makeatother

\author{\IEEEauthorblockN{Rui Zhang}
\IEEEauthorblockA{\textit{\small{Command \& Control Engineering College}} \\
\textit{Army Engineering University of PLA}\\
Nanjing, China \\
3959966@qq.com}
\and
\IEEEauthorblockN{Ziqi Wang}
\IEEEauthorblockA{\textit{\small{Command \& Control Engineering College    }}\\
\textit{Army Engineering University of PLA}\\
Nanjing, China \\
18071239797@163.com}
\and
\IEEEauthorblockN{Yang Li*\thanks{*Corresponding author.}}
\IEEEauthorblockA{\textit{\small{Command \& Control Engineering College}} \\
\textit{Army Engineering University of PLA}\\
Nanjing, China \\
solarleeon@outlook.com}
\linebreakand
\IEEEauthorblockN{Jiabao Wang}
\IEEEauthorblockA{\textit{\small{Command \& Control Engineering College}} \\
\textit{Army Engineering University of PLA}\\
Nanjing, China \\
jiabao\_1108@163.com}
\and
\IEEEauthorblockN{Zhiteng Wang}
\IEEEauthorblockA{\textit{\small{Communication Engineering College}} \\
\textit{Army Engineering University of PLA}\\
Nanjing, China \\
wangzhiteng168@163.com}
\and
}

\maketitle
\thispagestyle{fancy}
\fancyhead{}
\lhead{}
\lfoot{\copyright~ 978-1-6654-7726-0/22/\$31.00 ©2022 IEEE}
\cfoot{}
\rfoot{}

\begin{abstract}
Few-shot learning (FSL) is one of the significant and hard problems in the field of image classification. However, in contrast to the rapid development of the visible light dataset, the progress in SAR target image classification is much slower. The lack of unified benchmark is a key reason for this phenomenon, which may be severely overlooked by the current literature. The researchers of SAR target image classification always report their new results on their own datasets and experimental setup. It leads to inefficiency in result comparison and impedes the further progress of this area. Motivated by this observation, we propose a novel few-shot SAR image classification benchmark (FewSAR) to address this issue. FewSAR consists of an open-source Python code library of 15 classic methods in three categories for few-shot SAR image classification. It provides an accessible and customizable testbed for different few-shot SAR image classification task. To further understanding the performance of different few-shot methods, we establish evaluation protocols and conduct extensive experiments within the benchmark. By analyzing the quantitative results and runtime under the same setting, we observe that the accuracy of metric learning methods can achieve the best results. Meta-learning methods and fine-tuning methods perform poorly on few-shot SAR images, which is primarily due to the bias of existing datasets. We believe that FewSAR will open up a new avenue for future research and development, on real-world challenges at the intersection of SAR image classification and few-shot deep learning. We will provide our code for the proposed FewSAR at \url{https://github.com/solarlee/FewSAR}\href{https://github.com/solarlee/FewSAR}.

\end{abstract}


\section{Introduction}
Image classification is a fundamental problem in computer vision tasks \cite{b1,b2} and has achieved increasing attention over the past few decades \cite{b3,b4}. In recent years, although many experts and scholars have solved a lot of thorny problems, there are still many challenges. After years of research, the performance of image classification based on extensive training data has achieved great results \cite{AlexNet-NIPS-2012,ResNet-CVPR-2016}. But with fewer samples, the image classification task has become more and more difficult. Therefore, many scholars have turned their attention to exploring few-shot image classification in recent years \cite{b7}. On the visible light dataset, massive methods have achieved great performance \cite{b8}. And on certain datasets \cite{ImageNet-CVPR-2009}, many methods even got a higher accuracy than human.

SAR targets are images generated by synthetic aperture radar. Fig. \ref{fig1.1} shows the process of SAR target images classification. They have the characteristics of high resolution and a wide range of applications \cite{b10}. At the same time, the SAR target image has great application value in research directions such as natural disaster prediction and prevention, landform exploration, and ship identification \cite{b11}. However, due to the problem of signal blurring, it is easily affected by speckle noise for SAR target image processing \cite{DNN_for_SAR_classification-2016}. SAR target image classification depends on the design of the algorithm which leads to its poor robustness \cite{b13}. The problem of lack of samples is more prominent in SAR target images than in visible light images. Although some scholars have conducted some research in this field, they do not use a unified dataset. Besides, the lack of evaluation benchmarks makes it difficult to compare the final classification performance directly.

\begin{figure*}[hbtp]
\centerline{\includegraphics[width=1.0\linewidth, height=0.22\linewidth]{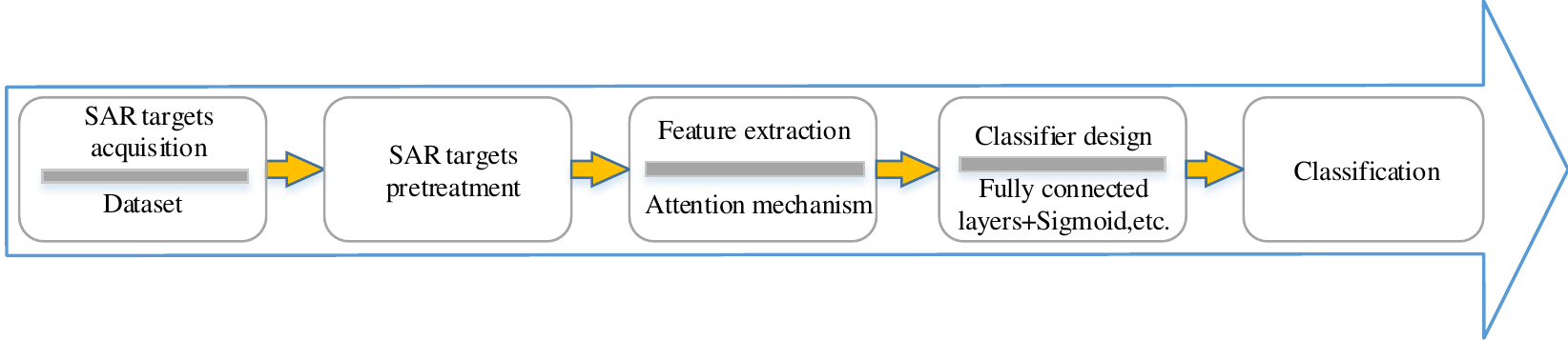}}
\caption{The process of SAR target images classification.}
\label{fig1.1}
\end{figure*}

Based on extensive researches in traditional few-shot methods, we propose a novel few-shot SAR image classification benchmark in this paper. We select 15 classic methods and divides them into three categories \cite{LibFewShot-aRxiv-2021}. Due to these methods have the problem of domain adaptation in SAR target recognition \cite{b15}, the effects between these 15 classic methods are still blank in this field. Therefore, we test 15 classic methods on the SAR target dataset, and propose a new evaluation benchmark to provide a reference standard for follow-up research.

In conclusion, our main contributions are as follows:
\begin{itemize}
    \item We propose a few-shot SAR image classification benchmark, named FewSAR, which is the first in the field of few-shot SAR image classification. It consists of 15 classic few-shot methods on the MSTAR dataset.
    
    \item We test the accuracy and runtime of the collected 15 algorithms on the MSTAR dataset. A lot of experimental results are evaluated and compared on the same experiment settings.

    \item Based on the benchmark proposed in this paper, we build a solid baseline on the MSTAR dataset using ATL\_Net method. We also provide an open-source codebase for easy reproduction and modification.
\end{itemize}

\section{Related Works}\label{sec2}

\subsection{Few-shot Learning}

Few-shot learning means to learn a task with very few training samples, which has been widely studied in recent years. In 2003, Li et al. \cite{b16} first proposed the concept of one-shot learning and hoped to solve it with a Bayesian framework \cite{b17}. With the rising development of deep learning, there has been better and more potential solutions to few-shot problems. At present, a variety of effective classification methods have been proposed for few-shot classification based on deep learning and convolutional neural network (CNN) \cite{AlexNet-NIPS-2012}.

The first is the fine-tuning method, which hopes to fine-tune the parameters of the network to achieve the classification of unknown new classes. Its idea comes from transfer learning \cite{b18,b19}. The Baseline and Baseline++ \cite{Baseline-ICLR-2019} proposed by Chen et al. are both divided into two stages. One of the stages is the training stage and the other is the fine-tuning stage. The SKD\_Model \cite{SKD_Model-arXiv-2020} hoped to learn a good feature embedding simply. It used a self-supervised auxiliary loss in the first stage and distillation process in the second stage. The RFS\_Model \cite{RFS_Model-2020} aimed to obtain an embedding model with better performance. On one hand, it optimizes the training process of the embedding model. On the other hand, it introduces the idea of knowledge distillation and further optimizes the feature extraction ability of the embedding model through multiple iterations.

\begin{figure*}[hbtp]
\centerline{\includegraphics[width=1.0\linewidth, height=0.22\linewidth]{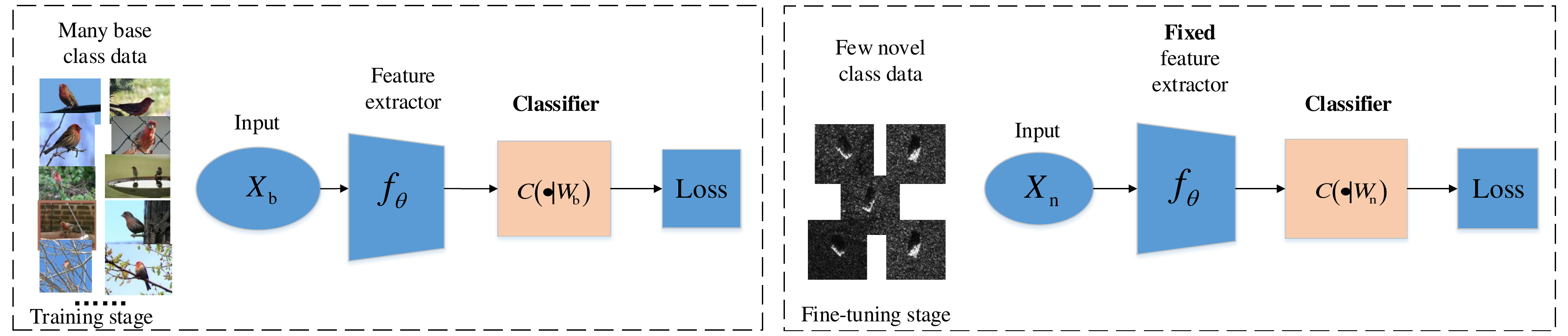}}
\caption{Structure diagram of few-shot image classification algorithm based on fine-tuning.}
\label{fig3.1}
\end{figure*}

The second is the meta-learning method, which hopes that machines can learn how to learn like humans. This method has been widely used in the field of machine learning, and it has become one of the effective methods to solve the problem of few-shot learning. The MAML \cite{MAML-ICML-2017} proposed by Finn et al. provided a model-agnostic algorithm which based on meta-learning. The Versa \cite{Versa-ICLR-2019} proposed by Gordon et al. is a flexible and versatile amortization network. It made the forward pass through the inference network replaces test-time optimization, amortizing the cost of inference and alleviating the need for second-order derivatives during training. The R2D2 \cite{R2D2-ICLR-2019} proposed used the main adaptation mechanism which based on some fast convergent methods for few-shot learning. The MTL \cite{MTL-CVPR-2019} introduced a method called meta-transfor learning (MTL). It learned to adapt a deep neural networks (DNNs) for few-shot learning tasks. The Leo \cite{Leo-ICLR-2019} proposed by Rusu et al. is a method called latent embedding optimization. It decouples the gradient-based adaptation procedure from the underlying high-dimensional space of model parameters. The ANIL \cite{ANIL-ICLR-2020} simplified the MAML which has almost no inner loop.

The last is the metric learning method, which hopes complete the classification task by measuring the distance between the sample of each class. The Proto\_Net \cite{Proto_Net-NeurIPS-2017} designed a prototypical networks, which computed distances to prototype representations of each class to complete classification tasks. The Feat \cite{Feat-CVPR-2020}  designed a few-shot embedding adaptation model. It adapted the instance embedding to the target classification task with a set-to-set function. The Relation Net (RN) \cite{RelationNet-CVPR-2018} is an end-to-end network composed of embedded modules and related modules. It computed relation scores between new samples of each new class and query images to achieve classification goals. The DN4 \cite{DN4-CVPR-2019} used a local descriptor based image-to-class measure instead of an image-level feature based measure. It designed a network called Deep Nearest Neighbor Neural Network which using a k-nearest neighbor search over the deep local descriptors of convolutional feature maps. The ATL\_Net \cite{ATL_Net-IJCAI-2020} focused on local representations (LRs). It designed a episodic attention mechanism which can explore and weight discriminative semantic patches to learn task-aware local representations. The CovaMNet \cite{CovaMNet-AAAI-2019}  designed Covariance Metric Networks, which is a novel end-to-end deep architecture. It measured consistency of distributions between query samples and new concepts by defining a new deep covariance metric. And CovaMNet trained an end-to-end network with an episodic training mechanism.

\subsection{Few-shot SAR Target Images Classification}
   SAR target images classification technology has always been a concern of many researchers, and the technology has been constantly updated and improved. Traditional technology is often driven by artificial cognition. For different tasks, feature extraction depends on artificially designed extractors, tuning, and optimizing through professional knowledge \cite{b10,DNN_for_SAR_classification-2016}. These methods have limitations, such as expensive costs, poor robustness, and difficult generalization. SAR target images classification method based on deep learning is increasingly becoming a feasible solution to this problem.

Early deep learning was restricted by data and computing power, and its development was relatively slow. With the advent of the information age and the application of GPU in the field of deep learning, deep learning has developed rapidly. However, SAR target images have the characteristics of a few existing samples, and it is necessary to solve the classification problem through few-shot learning. The DKTS-N \cite{DKTS-N-IEEE-2022} proposed by Zhang et al. in 2022 designed a domain knowledge-powered two-stream deep network. It introduced domain knowledge in combination with datasets to assist the classification task of SAR target vehicle images. The SCAN \cite{SCAN-IEEE-2022} designed a scattering characteristics analysis network. It used a scattering extraction module (SEM) to adapt to the scattering properties of SAR target images. Although some scholars have conducted some research in the field of SAR target images classification, they have not been able to compare the classification effects under the general standard.

\begin{figure*}[hbtp]
\centerline{\includegraphics[width=0.70\linewidth, height=0.30\linewidth]{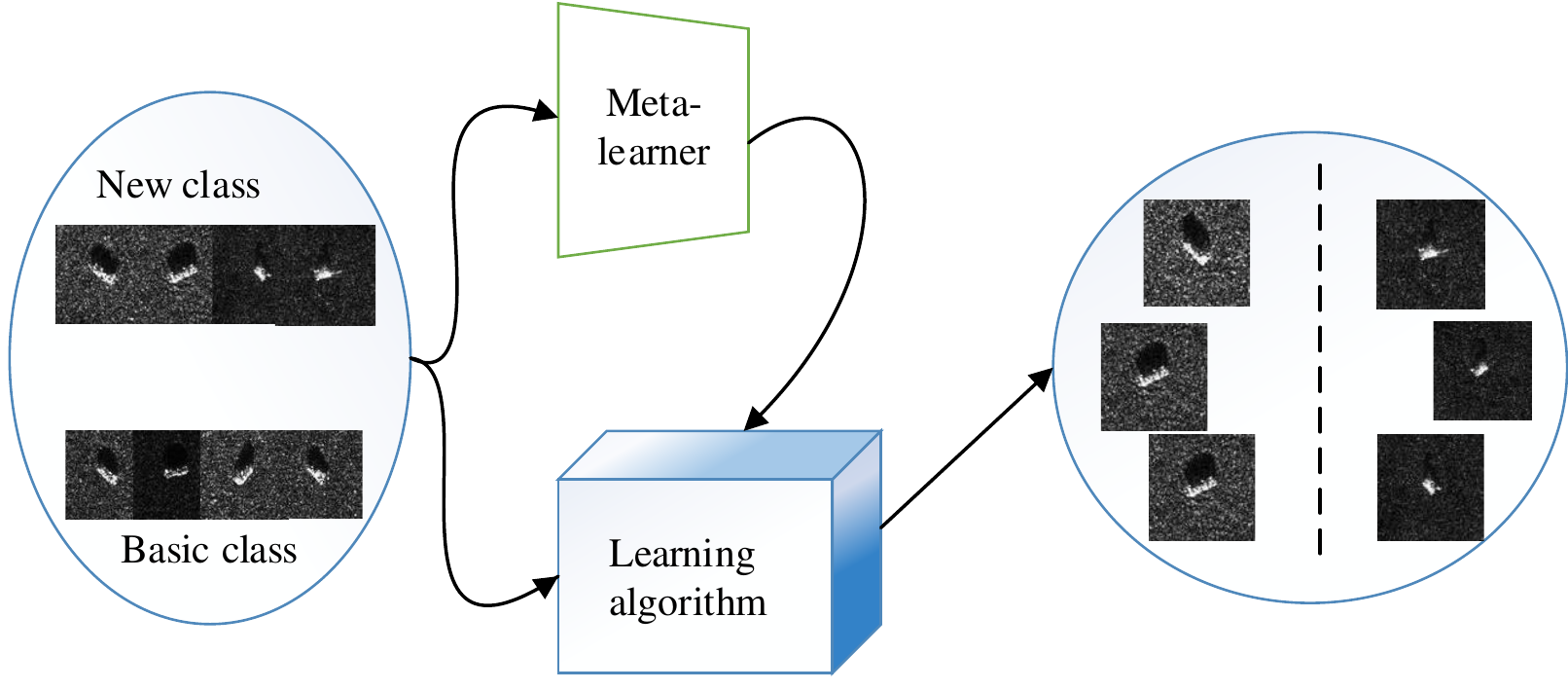}}
\caption{Structure diagram of few-shot image classification algorithm based on meta-learning.}
\label{fig3.2}
\end{figure*}

\section{Few-shot Learning for SAR Target Images Benchmark}\label{sec3}

In recent years, many classic methods have been proposed to solve few-shot learning problems in image classification \cite{A_Comprehensive_Survey_of_Few-shot_Learning}. We test 15 classic methods in a unified framework with the help of a comprehensive few-shot algorithm library, named Libfewshot \cite{LibFewShot-aRxiv-2021}. Libfewshot integrates the classic few-shot learning methods into one framework. By calling different configuration files, different methods can be quickly switched, and it is also convenient for us to fairly compare the performance of various methods on the MSTAR dataset \cite{MSTAR_Dataset}. To test SAR target images classification benchmark, we selected a total of 15 methods in three categories: the fine-tuning method, the meta-learning method, and the metric learning method.

\subsection{Fine-tuning Methods}
There are three methods based on fine-tuning, named Baseline \cite{Baseline-ICLR-2019}, Baseline++ \cite{Baseline-ICLR-2019}, and SKD\_Model \cite{SKD_Model-arXiv-2020}. The basic idea of this type of method is to pre-train model parameters on a large-scale dataset, and then fine-tune the parameters of the top layers or fully-connected layers of the neural network on a small-sample dataset. This is a traditional and simple method, but it is prone to overfitting problems. In Baseline++, ${X_b}$ represents abundant base data, and ${X_n}$ represents a handful of novel class data which has been labeled. As we can see in Fig. \ref{fig3.1}, in the training stage, there is a weight matrix of ${W_b} \in {\mathbb{R}^{d \times c}}$. And there is a ${W_n}$ in the fine-tuning stage. Then writing the weight matrix ${W_b}$ as $\left[ {{w_1},{w_2},...,{w_c}} \right]$, where a $d$-dimensional weight vector in each class. Wei-Yu Chen et al. \cite{Baseline-ICLR-2019} computes its cosine similarity to each weight vector $\left[ {{w_1},{w_2},...,{w_c}} \right]$ and all classes similarity scores $\left[ {{s_{i,1}},{s_{i,2}},...,{s_{i,c}}} \right]$ will be obtained for a input feature ${f_\theta }\left( {{x_i}} \right)$, where ${x_i} \in {X_b}$. We can use Eq. (\ref{eq1}) to get the similarity scores.
\begin{equation}
{S_{i,j}} = {f_\theta }{\left( {{x_i}} \right)^\top }{w_j}/\left\| {\left. {{f_\theta }\left( {{x_i}} \right)} \right\|} \right.\left\| {\left. {{w_j}} \right\|} \right.\ 
\label{eq1}
\end{equation}

\subsection{Meta-learning Methods}
There are six methods based on meta-learning, named MAML \cite{MAML-ICML-2017}, Versa \cite{Versa-ICLR-2019}, R2D2 \cite{R2D2-ICLR-2019}, MTL \cite{MTL-CVPR-2019}, Leo \cite{Leo-ICLR-2019}, and ANIL \cite{ANIL-ICLR-2020}. The basic idea of this type of method is shown in Fig. \ref{fig3.2}. Meta-learning method needs to train a meta-learner, so the meta-learner can guide the classification task of new classes after gaining learning experience. MAML\cite{MAML-ICML-2017} is a representative meta-learning method. It is a model-agnostic meta-learning method combined with any deep learning model. It does not introduce other learning parameters for the original model but uses the idea of gradient optimization to achieve parameter update. MAML defines the optimization problem as a two-layer optimization problem. The inner layer optimizes the task-related parameters, and the outer layer optimizes the task-independent parameters. Both of them use the gradient descent method. 

As shown in Eq. (\ref{eq2}), $T$ represents the sampled task, $f$ represents the original model, $\alpha$ represents the learning rate of the inner layer optimization, and $\beta$ represents the learning rate of the outer layer optimization. Then, the objective function $G\left({\rm{x}} \right)$ is defined as Eq. (\ref{eq2}):

\begin{equation}
G\left( {\rm{x}} \right){\rm{ = }}\mathop {\min }\limits_\theta  \sum\limits_{{T_i} \sim p\left( T \right)} {{L_{{T_i}}}\left( {{f_\theta } - \alpha {\nabla _\theta }{L_{{T_i}}}\left( {{f_\theta }} \right)} \right)} \label{eq2}
\end{equation}

The objective function has been updated with the stochastic gradient descent algorithm to complete the inner layer optimization as Eq. (\ref{eq3}):

\begin{equation}
{\theta _i}^\prime  = \theta  - \alpha {\nabla _\theta }{L_{{T_i}}}\left( {{f_\theta }} \right)
 \label{eq3}
\end{equation}

Then MAML \cite{MAML-ICML-2017} uses the same optimization algorithm to update the parameters of the outer layer as Eq. (\ref{eq4}):

\begin{equation}
\theta  \leftarrow \theta  - \beta {\nabla _\theta }\sum\limits_{{T_i} \sim p\left( T \right)} {{L_{{T_i}}}\left( {{f_{{\theta _i}^\prime }}} \right)}
 \label{eq4}
\end{equation}

\begin{figure*}[hbtp]
\centerline{\includegraphics[width=0.94\linewidth, height=0.30\linewidth]{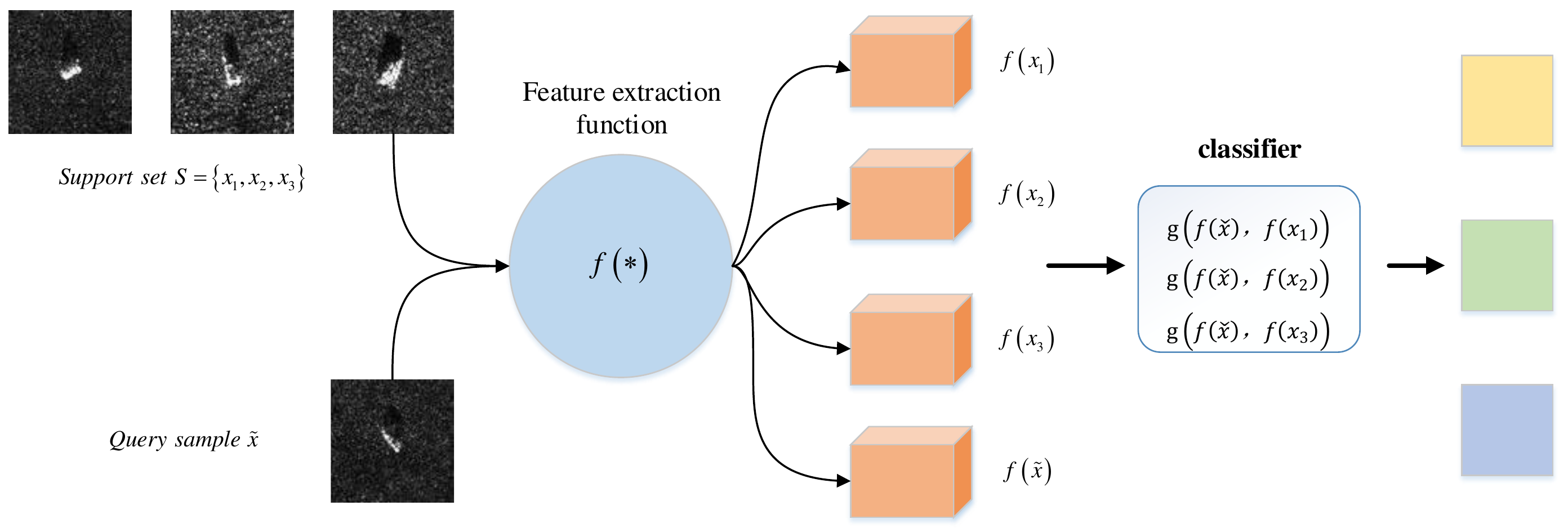}}
\caption{Structure diagram of few-shot image classification algorithm based on metric learning.}
\label{fig3.3}
\end{figure*}

\subsection{Metric learning Methods}
There are six methods based on metric learning, named Proto\_Net \cite{Proto_Net-NeurIPS-2017}, Feat \cite{Feat-CVPR-2020}, Relation Net \cite{RelationNet-CVPR-2018}, DN4 \cite{DN4-CVPR-2019}, ATL\_Net \cite{ATL_Net-IJCAI-2020}, and CovaMNet \cite{CovaMNet-AAAI-2019}. The basic idea of this kind of method is to change the original classification method to measure the distance between the query sample and the category center. Taking a 5-way 1-shot task as an example, we can see from Fig. \ref{fig3.3} that the support set samples and 1 query sample are sent to the feature extractor to get the feature extraction. Then the output query sample features are the support set sample features, respectively. Then, the function named $g\left( x \right)$ obtains the result after implementing the similarity measure. ATL\_Net \cite{ATL_Net-IJCAI-2020} proposed an episode attention mechanism that aims to learn task-aware local representations (LRs) for few-shot learning. It realizes the selection of the most informative relationship by learning different thresholds for different patches. Specifically, the author used a multi-layer perceptron (MLP) to predict the threshold for each local representation of the query image adaptively. It can be expressed by the Eq. (\ref{eq5}):

\begin{equation}
{{V_{\rm{c}}}{\rm{ = }}\sigma \left( {{F_\theta }\left( {L_{\rm{i}}^{\rm{q}}} \right)} \right)} 
\label{eq5}
\end{equation}

In Eq. (\ref{eq5}), ${L^q}$ represents the $HW$-dimensional query LRs and $i \in \left\{ {1,...,HW} \right\}$. ${F_\theta }$ takes the LRs of the query set as input and outputs a threshold ${V_{\rm{c}}}$. Through this threshold, ATL\_Net can select and weight key patches to achieve better classification results.


\section{Experiments}\label{sec4}
This section shows experimental results on the MSTAR dataset \cite{MSTAR_Dataset}. Section \ref{tex4.1}  introduces the details of the experiment from three aspects. Section \ref{tex4.2} presents a quantitative performance that uses accuracy as the standard. Section \ref{tex4.3} analyzes all methods from a runtime perspective. 

\subsection{Experimental Details}\label{tex4.1}
\textbf{Dataset.} MSTAR \cite{MSTAR_Dataset} is a synthetic aperture radar dataset, which is commonly used for SAR target images classification. It consists of 10 classes and more than 200 images per class. The MSTAR public targets chips of the dataset mainly include three types of target image data: T72, BMP2, and BTR70, namely three types of target recognition problem data. There are different methods of targets in various categories. The differences in targets in different methods of the same kind are reflected in equipment, but the difference of all kinds is small. This paper uses the MSTAR dataset to test the performance of many classic few-shot learning methods for SAR targets classification.

\textbf{Network architecture.}  We uniformly choose Conv64F as the backbone to compare 15 classic few-shot learning methods. Due to more and more scholars pay attention to the local features and local representations of images when solving the few-shot problems \cite{ATL_Net-IJCAI-2020,DN4-CVPR-2019}. Compared with Transformer \cite{Transformer-2017}, the CNN moves a fixed step size through the convolution kernel, which can better extract the local features of the images.

\textbf{Training and testing detail.} We use Python3.6 to implement our experiments, and all the images in MSTAR are resized to 84*84, including 2S1, BRDM\_2, BTR 70, D7, and T72, etc. We divide the MSTAR dataset into a training set and a test set, where the training set and the test set each have 5 categories of images. Since few-shot learning aims to learn new categories, the categories in the two parts cannot be the same. We collect 15 query images per class in each episode to test the accuracy of the method. For example, under the 5-way 1-shot setting, each task has 5 support images and 75 query images, which selects from the corresponding partitioned dataset randomly. We use an Adam optimizer \cite{Adam-ICLR-2015}, which initial learning rate is set to 0.001 with a cross-entropy loss to train our network. All experiments were run on a computer with an NVIDIA RTX1070 GPU and i7-7820HK CPU.

\begin{table*}[hbtp]
\renewcommand{\arraystretch}{1.2} 
\caption{Classification accuracy of all methods. The best value in each categories is highlighted in {\color[HTML]{C00000} red}}
\centering
\begin{tabular}{cccccc}
\hline
Model        & Venue        & Category   & Backbone & 5-way 1-shot                   & 5-way 5-shot                   \\ \hline
Baseline\cite{Baseline-ICLR-2019}     & ICLR 2019    & fine-tuning & Conv64F  & 54.44\%                        & 83.13\%                        \\
Baseline++\cite{Baseline-ICLR-2019}   & ICLR 2019    & fine-tuning & Conv64F  & {\color[HTML]{FE0000} 59.98\%} & {\color[HTML]{FE0000} 86.37\%} \\
SKD\_Model\cite{SKD_Model-arXiv-2020}   & arXiv 2020   & fine-tuning & Conv64F  & 57.92\%                        & 78.39\%                        \\ \hline
MAML\cite{MAML-ICML-2017}         & ICML 2017    & meta       & Conv64F  & 19.87\%                        & 60.67\%                        \\
Versa\cite{Versa-ICLR-2019}        & ICLR 2019 & meta       & Conv64F  & {\color[HTML]{FE0000} 66.96\%} & 68.01\%                        \\
R2D2\cite{R2D2-ICLR-2019}         & ICLR 2019    & meta       & Conv64F  & 63.99\%                        & {\color[HTML]{FE0000} 68.88\%} \\
MTL\cite{MTL-CVPR-2019}          & CVPR 2019    & meta       & Conv64F  & 18.13\%                        & 47.07\%                        \\
Leo\cite{Leo-ICLR-2019}          & ICLR 2019    & meta       & Conv64F  & 36.00\%                        & 44.00\%                        \\
ANIL\cite{ANIL-ICLR-2020}         & ICLR 2020    & meta       & Conv64F  & 20.99\%                        & 61.91\%                        \\ \hline
Proto\_Net\cite{Proto_Net-NeurIPS-2017}   & NeurIPS 2017 & metric     & Conv64F  & 39.68\%                        & 42.72\%                        \\
Feat\cite{Feat-CVPR-2020}         & CVPR 2020    & metric     & Conv64F  & 46.11\%                        & 56.36\%                        \\
Relation Net\cite{RelationNet-CVPR-2018} & CVPR 2018    & metric     & Conv64F  & 64.84\%                        & 77.51\%                        \\
DN4\cite{DN4-CVPR-2019}          & CVPR 2019    & metric     & Conv64F  & 67.37\%                        & 85.15\%                        \\
ATL\_Net\cite{ATL_Net-IJCAI-2020}     & IJCAI 2020   & metric     & Conv64F  &{\color[HTML]{FE0000} 72.03\%} &{\color[HTML]{FE0000} 88.81\%}                        \\
CovaMNet\cite{CovaMNet-AAAI-2019}     & AAAI 2019    & metric     & Conv64F  & 58.75\%                        & 45.75\%                        \\ \hline
\end{tabular}
\label{tab4.1}
\end{table*}

\subsection{Quantitative Performance Comparison}\label{tex4.2}
Table \ref{tab4.1} shows the average accuracy of the 15 methods tested in this paper on the MSTAR dataset. And all methods in three categories are tested under the 5-way 1-shot and 5-way 5-shot tasks. We can see that the methods based on the metric learning achieve the best performance than the other methods, and the meta-learning methods perform poorly overall. Due to the meta-learning methods rely on the prior knowledge learned by the meta-learner, and the existing SAR target dataset has few training samples. They cannot support the learning of the meta-learner very well, which results in the decline of the overall accuracy. In addition, the fine-tuning methods make use of the initial parameters pre-trained on visible light images, and domain knowledge is not well utilized when transferring to SAR target images across domains.  

Besides, ATL\_Net \cite{ATL_Net-IJCAI-2020} and DN4 \cite{DN4-CVPR-2019} achieve optimal and sub-optimal results respectively among the methods of metric learning. Combined with the analysis of specific methods, we find that both methods focus on local representations. The characteristics of SAR target images determine to focus on their local features, which is beneficial to the realization of classification work. Finally, ATL\_Net \cite{ATL_Net-IJCAI-2020} achieves the best classification accuracy in both 1-shot and 5-shot tasks among all methods. ATL\_Net \cite{ATL_Net-IJCAI-2020} also designs a novel episodic attention mechanism by exploring and weighting discriminative semantic patches across the entire task for few-shot learning.

\begin{table*}[hbtp]               
\renewcommand{\arraystretch}{1.2} 
\caption{Runtime of all methods}
\centering
\begin{tabular}{cccccc}
\hline
Model        & Venue        & Category   & Backbone & 5-way1-shot (min) & 5-way 5-shot (min) \\ \hline
Baseline\cite{Baseline-ICLR-2019}     & ICLR 2019    & fine-tuning & Conv64F  & 0.9              & 2.18              \\
Baseline++\cite{Baseline-ICLR-2019}   & ICLR 2019    & fine-tuning & Conv64F  & 5.85             & 15.27             \\
SKD\_Model\cite{SKD_Model-arXiv-2020}   & arXiv 2020   & fine-tuning & Conv64F  & 0.48             & 0.52              \\ \hline
MAML\cite{MAML-ICML-2017}         & ICML 2017    & meta       & Conv64F  & 0.1              & 0.18              \\
Versa\cite{Versa-ICLR-2019}        & ICLR 2019 & meta       & Conv64F  & 0.72             & 0.76              \\
R2D2\cite{R2D2-ICLR-2019}         & ICLR 2019    & meta       & Conv64F  & 0.58             & 0.68              \\
MTL\cite{MTL-CVPR-2019}          & CVPR 2019    & meta       & Conv64F  & 0.08             & 0.1               \\
Leo\cite{Leo-ICLR-2019}          & ICLR 2019    & meta       & Conv64F  & 0.08             & 0.1               \\
ANIL\cite{ANIL-ICLR-2020}         & ICLR 2020    & meta       & Conv64F  & 0.62             & 0.84              \\ \hline
Proto\_Net\cite{Proto_Net-NeurIPS-2017}   & NeurIPS 2017 & metric     & Conv64F  & 0.46             & 0.52              \\
Feat\cite{Feat-CVPR-2020}         & CVPR 2020    & metric     & Conv64F  & 0.6              & 0.74              \\
Relation Net\cite{RelationNet-CVPR-2018} & CVPR 2018    & metric     & Conv64F  & 0.64             & 0.78              \\
DN4\cite{DN4-CVPR-2019}          & CVPR 2019    & metric     & Conv64F  & 0.64             & 0.8               \\
ATL\_Net\cite{ATL_Net-IJCAI-2020}     & IJCAI 2020   & metric     & Conv64F  & 0.63             & 0.76              \\
CovaMNet\cite{CovaMNet-AAAI-2019}     & AAAI 2019    & metric     & Conv64F  & 0.62             & 0.76              \\ \hline
\end{tabular}
\label{tab4.2}
\end{table*}

\subsection{Runtime Comparison}\label{tex4.3}
Table \ref{tab4.2} shows the runtime of all methods. We analyze the running efficiency of various methods to achieve a comprehensive evaluation of each method. We trained the same number of epochs for each method, so we took the average runtime for comparison. Overall, the fine-tuning methods have the longest runtime in the three categories. We believe the reason for this phenomenon is that it takes a long time to fine-tune the baseline model by learning new knowledge. The meta-learning methods have the shortest runtime, but the classification accuracy is poor. The reason may be that the MSTAR dataset has less data than the visible light dataset. During the training stage, the meta-learner fails to learn enough knowledge, which results in a generally low final accuracy and a corresponding reduction in training time. 

The metric learning methods have a short runtime and can achieve better classification accuracy. It is the most suitable method for SAR target image classification at this stage. This kind of method cleverly uses the distance between the class and the center of the class to discriminate the feature similarity, which simplifies the way of discriminating the similarity. In particular, the two methods of DN4 \cite{DN4-CVPR-2019} and ATL\_Net \cite{ATL_Net-IJCAI-2020} are comparable in accuracy to the optimal value of the fine-tuning, but the runtime is greatly shortened compared with Baseline \cite{Baseline-ICLR-2019} and Baseline++ \cite{Baseline-ICLR-2019}. It is worthy of our further exploration and research.

\section{Concluding Remarks}\label{sec5}
This paper proposes a comprehensive evaluation benchmark for SAR target images classification based on few-shot learning. Meanwhile, 15 classic methods of few-shot learning were compared in three categories based on the general MSTAR dataset. To the best of our knowledge, this is the first few-shot learning benchmark for SAR target images classification to date. This benchmark can help us clearly understand the performance of various classic methods in the SAR targets field. Combined with the runtime, we can also comprehensively select methods with faster performance, and use this as a baseline for optimization and improvement.

In our future work, we will continue to complement our benchmark by increasing the code library with new SAR targets classification datasets and the latest methods. At the same time, we also hope to explore the evaluation criteria of SAR targets classification based on few-shot learning and propose more evaluation metrics reasonably.

\section*{Acknowledgments}
This work has been supported by the Natural Science Foundation of Jiangsu Province (BK20200581), and in part by the Basic Frontier Science Innovation Project of Army Engineering University of PLA.

\balance
\bibliographystyle{IEEEtran}
\bibliography{refs}

\end{document}